\documentclass{article}

\usepackage[preprint]{neurips_2025}

\usepackage[utf8]{inputenc} 
\usepackage[T1]{fontenc}    
\usepackage{hyperref}       
\hypersetup{
    hidelinks,
    colorlinks=false,
    linkcolor=black,
    citecolor=black
}
\usepackage{url}            
\usepackage{booktabs}       
\usepackage{amsfonts}       
\usepackage{nicefrac}       
\usepackage{microtype}      
\usepackage{xcolor}         
\usepackage{natbib}         
\usepackage{multirow}
\usepackage{fontawesome}
\usepackage{float}
\usepackage{makecell}
\usepackage{graphicx}
\usepackage{subfigure}
\usepackage{booktabs}
\usepackage[table]{xcolor}

\title{LiveAgentBench: Comprehensive Benchmarking of Agentic Systems Across 104 Real-World Challenges}

\makeatletter
\@submissionfalse  
\makeatother

\author{%
  Hao Li \\
  Ant Group\\
  \texttt{lh460759@antgroup.com} \\
  \And
  Huan Wang \\
  Ant Group \\
  \texttt{huan.wh@antgroup.com} \\
  \AND
  Jinjie Gu \\
  Ant Group \\
  \texttt{jinjie.gujj@antgroup.com} \\
  \And
  Wenjie Wang \\
  Ant Group \\
  \texttt{xiaowen.wwj@antgroup.com} \\
  \AND
  Chenyi Zhuang \\
  Ant Group \\
  \texttt{chenyi.zcy@antgroup.com} \\
  \And
  Sikang Bian \\
  Ant Group \\
  \texttt{biansikang.bsk@antgroup.com} \\
}

\begin{document}

\maketitle

\begin{abstract}
As large language models grow more capable, general AI agents have become increasingly prevalent in practical applications. However, existing benchmarks face significant limitations, failing to represent real-world user tasks accurately. To address this gap, we present LiveAgentBench, a comprehensive benchmark with 104 scenarios that reflect real user requirements. It is constructed from publicly sourced questions on social media and real-world products. Central to our approach is the Social Perception-Driven Data Generation (SPDG) method, a novel process we developed to ensure each question's real-world relevance, task complexity, and result verifiability. We evaluate various models, frameworks, and commercial products using LiveAgentBench, revealing their practical performance and identifying areas for improvement. This release includes 374 tasks, with 125 for validation and 249 for testing. The SPDG process enables continuous updates with fresh queries from real-world interactions. 
\end{abstract}

\section{Introduction}

In recent years, Artificial Intelligence has gradually shifted from single-task processing to a paradigm of decision execution for complex tasks, driven by the development of reasoning Large Language Models (LLMs) and autonomous agent technologies. Reasoning models represented by DeepSeek-R1\citep{DeepSeek}, OpenAI o3\citep{OpenAI_O3}, have achieved good scores in math, code generation and knowledge benchmarks, which include GPQA\citep{GPQA}, LiveCodeBench\cite{LiveCodeBench}, Codeforces\footnote{https://codeforces.com/} and AIME 2024\citep{AIME2024}. However, the evaluation scope of these benchmarks is relatively homogeneous, and there is still a big gap compared with complex tasks in the human world. Based on the powerful reasoning and decision-making capability, many companies have gradually released their deep research agents or applications to improve the ability of LLMs to think and answer complex questions, such as OpenAI Deep Research\citep{OpenAI_Deep_Research}, Perplexity Deep Research\citep{Perplexity_Deep_Research} and Gemini Deep Research\citep{Gemini_Deep_Research}. These deep research agents can decompose complex research tasks, perform multiple-step searches, read a large amount of information, and integrate materials to generate a comprehensive and in-depth report. Meanwhile, in the field of intelligence, many autonomous agents that can handle complex tasks have emerged, such as Manus\citep{Manus}, MetaGPT\citep{MetaGPT}, and AWorld\citep{AWorld2025}. Different from deep research, these agents not only have strong thinking ability, but also can complete real-world tasks autonomously.

\begin{figure}[h]
  \centering
  \includegraphics[width=1\textwidth]{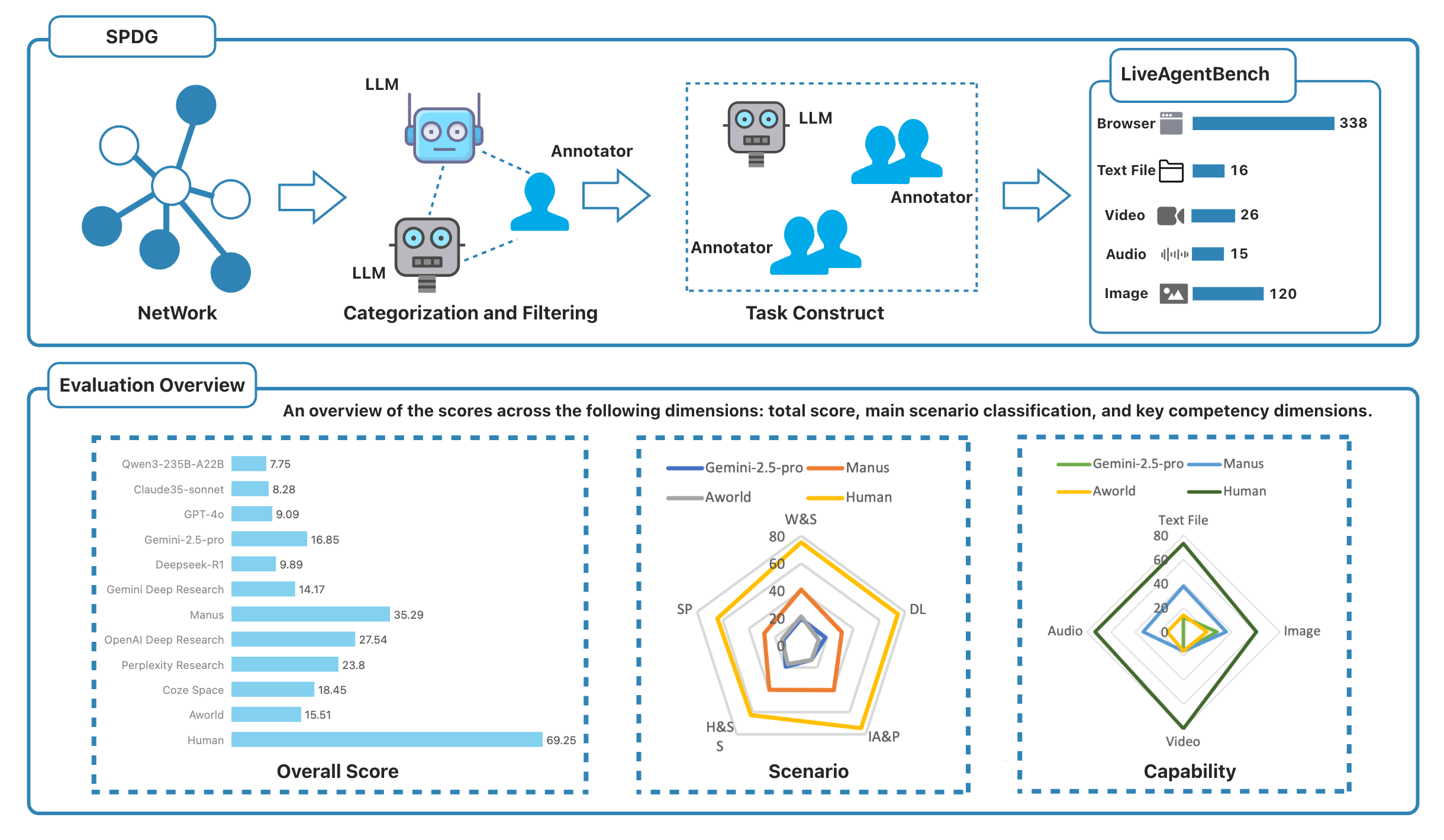}
  \caption{An overview of LiveAgentBench, introducing the construction process of the evaluation dataset from real user cases. It is accompanied by the summary results of LiveAgentBench. "W\&S" represents Work and Study, "DL" represents Daily Life, "IA\&P" represents Information Access and Processing, "H\&SS" represents Humanities and Social Science, and "SP" represents Social Production.}
\end{figure}

As all kinds of autonomous agents begin to solve real-world problems, a benchmark that can comprehensively evaluate real-world tasks is crucial. First of all, unlike the past benchmarks such as MMLU-Pro\citep{MMLU-Pro}, AGIEval\citep{AGI-Eval}, GSM8k\citep{GSM8k}, which focus on a single capability, for real-world complex tasks, agents should have multimodal processing, tool use, and strong reasoning capabilities at the same time. Benchmarks like GAIA\citep{GAIA} and AgentBench\citep{AgentBench} provide several real-world evaluation tasks for the reasoning, multimodal, and web browsing capabilities of agents. However, their scope is still insufficient for human real-world tasks, such as phone use and video comprehension, which are high-frequency scenarios in human daily life. On the other hand, regular maintenance and updates of the dataset are also necessary. Since part of the evaluation data, like browser operations in the dataset, has high uncertainty. Changes in webpage information will directly cause the dataset to be unavailable. As a result, the robustness and accuracy of the evaluation results will be affected. In addition, LLMs are usually trained with massive unrecognisable corpora, and current datasets are at high risk of contamination as they may be included in these training data\citep{LiveCodeBench}. Therefore, by updating the dataset regularly, unconvincing evaluation results due to the contaminated dataset can be avoided.

Motivated by these issues, we propose LiveAgentBench, a dynamically updated benchmark for comprehensive agent evaluation. LiveAgentBench covers 104 daily real-world scenarios by collecting real users' questions from different internet platforms and social media. In these scenarios, agents should have multiple capabilities such as browser operation, file operation, Android/IOS system operation, audio and video comprehension. Queries in different scenarios are strictly screened to ensure that they are suitable for evaluating the distinct capabilities of the agent, taking into account the difficulty of the questions and the dimensions examined. What's more, the ground truth of all questions is collected through double-blind labelling, and a third person is introduced to review the answers if the results of two people are inconsistent, which ensures that all the answers are correct and confident. When processing the evaluation step, we use the zero-shot prompt and extract the answers from agents' responses and compare them with the ground truth.

Users are always willing to share and communicate with each other in open communities or platforms on the Internet when they encounter problems. In order to create more realistic and relevant questions, we collected a large amount of data from different websites, Apps, and videos through automated and manual collection methods. Moreover, to ensure that the tasks are more challenging, easy to verify, and can assess different capabilities of agents in real-world scenarios, we carefully filtered the corpus and obtained 104 scenario categories and hundreds of seed questions. Since most real users' questions are still relatively open-ended, they are difficult to evaluate due to the lack of a fixed answer. Therefore, we involve manual labelling to modify the questions appropriately, making the answer fixed without changing the ability examined in the question. During this process, dozens of annotators collaborated to complete data generation, and we refined the entire data generation process into a sustainable and standard workflow, named Social Perception-Driven Data Generation (SPDG), which is helpful to ensure that further data supplementation and updates can be executed efficiently.

\begin{table}[t]\small
    \caption{Differences between LiveAgentBench and other benchmarks}
    \label{Differences}
    \centering
    \renewcommand\arraystretch{1.3}
    \tabcolsep=0.03cm
    \begin{tabular}{cccccc}
        \toprule
        \multirow{2}{*}{\makecell[c]{\\ Categories}} & \multirow{2}{*}{\makecell[c]{\\ Subcategories}} & \multicolumn{4}{c}{Benchmarks} \\
        \cmidrule(r){3-6}
        & & \makecell[c]{GAIA \\ \citep{GAIA}} & \makecell{AgentBench \\ \citep{AgentBench}} & \makecell[c]{API-Bank \\ \citep{API_Bank}} & LiveAgentBench \\
        \midrule
        \multirow{5}{*}{\makecell[c]{\\ Tool Use}} & Browser & \color{teal}\faCheckCircle & \color{teal}\faCheckCircle &                  \color{red}\faTimesCircle & \color{teal}\faCheckCircle \\
                                  & Text File & \color{teal}\faCheckCircle & \color{red}\faTimesCircle &                  \color{red}\faTimesCircle & \color{teal}\faCheckCircle \\
                                  & Android/IOS OS & \color{red}\faTimesCircle & \color{red}\faTimesCircle &                  \color{red}\faTimesCircle & \color{teal}\faCheckCircle \\
                                  & Audio & \color{teal}\faCheckCircle & \color{red}\faTimesCircle &                  \color{red}\faTimesCircle & \color{teal}\faCheckCircle \\
                                  & Video & \color{teal}\faCheckCircle & \color{red}\faTimesCircle &                  \color{red}\faTimesCircle & \color{teal}\faCheckCircle \\
                                  & Image & \color{teal}\faCheckCircle & \color{teal}\faCheckCircle &                  \color{red}\faTimesCircle & \color{teal}\faCheckCircle \\
        \midrule
        \multirow{2}{*}{Diversity} & Real-World Scenarios & \color{red}\faTimesCircle & \color{red}\faTimesCircle &                  \color{red}\faTimesCircle & \color{teal}\faCheckCircle \\
                                   & Real-World Cases & \color{teal}\faCheckCircle & \color{teal}\faCheckCircle &                  \color{red}\faTimesCircle & \color{teal}\faCheckCircle \\
        \midrule
        Regular Updates & Dataset & \color{red}\faTimesCircle & \color{red}\faTimesCircle &                  \color{red}\faTimesCircle & \color{teal}\faCheckCircle \\
        \bottomrule
    \end{tabular}
\end{table}

We evaluated current mainstream open-source and closed-source autonomous agents and LLMs such as Manus\citep{Manus}, Perplexity Deep Research\citep{Perplexity_Deep_Research}, GPT-4o\citep{GPT-4o}, Qwen3-235B\citep{Qwen3} on LiveAgentBench. Our key results are as follows: \textbf{1)} On LiveAgentBench, all of these products performed poorly. Even the best performing product (Manus) has a success rate of only 35.29\%, while humans can achieve a rate of 69.25\%. \textbf{2)} On average, agents with inner tools perform 56.51\% better than LLMs on LiveAgentBench, and these improvements mainly come from abundant tools to use, and have a certain level of task planning and decision-making capabilities. However, the stability of tools has a greater impact on agents' performance. \textbf{3)} Additionally, the lack of environmental background knowledge will prevent agents from obtaining the information when entering an unfamiliar website. \textbf{4)} Compared to other agents, the success rate between AWorld and other agents is around 8.34\%, which is mainly due to the stability. During the experiment using AWorld, approximately 11.76\% of tasks failed to execute due to the instability.

\section{Related Work}
\subsection{Autonomous Agents}

Recently, with the great development of large language models, the capabilities of LLMs' decision-making and reasoning have been significantly improved, which is beneficial for autonomous agents\citep{Abilities_of_LLMs,ReAct,Voyager}. Many works have improved the problem-solving abilities of LLMs through multiple agents\citep{MetaGPT,Survey_on_agents,A_Task-Solving_Agent,Improving_Factuality_and_Reasoning_in_LLMs}, and several multi-agent frameworks have been proposed as well. For example, AWorld\citep{AWorld2025}, a multi-agent execution MCP server, bridges the gap between theoretical  Multi-Agent System capabilities. MetaGPT\citep{MetaGPT}  is a meta-programming framework for LLM-based multi-agent systems. Despite the continuous emergence of various types of agents, there has been a relative lack of suitable and comprehensive benchmarks.

\subsection{Evaluating Agents}

Many benchmarks have been proposed to evaluate different capabilities of autonomous agents, while most of them only focus on a specific field. PlanBench\citep{PlanBench} or Natural Plan\citep{NATURAL_PLAN}, for example, are designed to assess the performance of LLMs in planning and reasoning. Moreover, several benchmarks provide different API tools to evaluate LLMs' capabilities in calling APIs\citep{API_Bank,T-Eval,ToolLLM,Gorilla,StableToolBench,ComplexFuncBench}. Besides, studies like AndroidWorld\citep{AndroidWorld} and OSWorld\citep{OSWorld} are designed to evaluate the capabilities of LLMs in operating different systems, including Android, Windows, MacOS. GAIA\citep{GAIA} and AgentBench\citep{AgentBench} provide a more generalised dataset for autonomous agents. However, there are still some gaps with real-world scenarios. Our study focuses on expanding real-world datasets and proposes a sustainable and standardised data production process.

\section{LiveAgentBench}
\subsection{Overview}

LiveAgentBench is an open source benchmark for evaluating autonomous agents, and it follows the three principles of realistic relevance, challenge, and ease of validation. These three principles are reflected in the following key aspects.

\begin{figure}[h]
  \centering
  \includegraphics[width=1\textwidth]{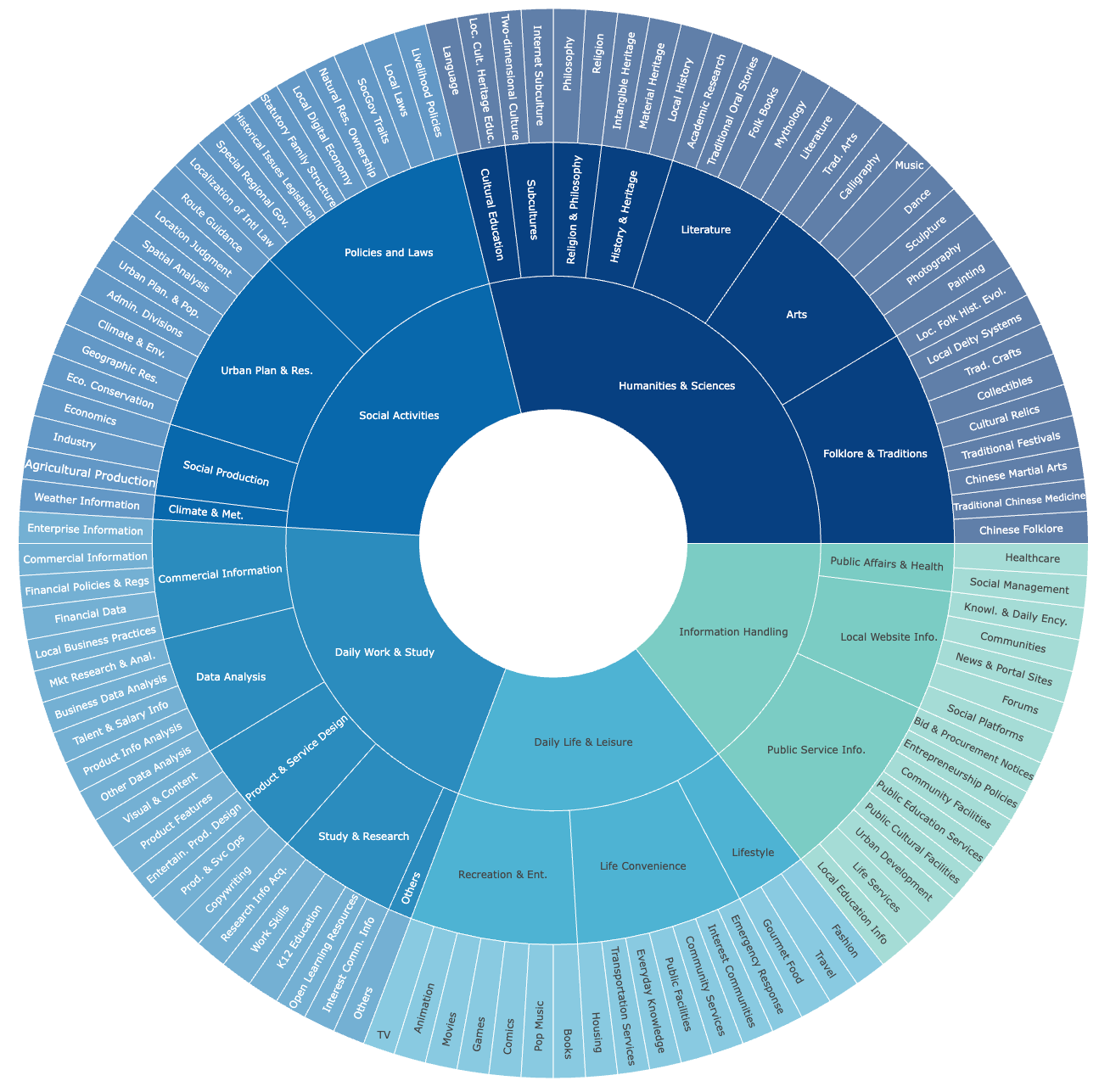}
  \caption{104 Real-World Challenges in LiveAgentBench.}
\end{figure}

\textbf{Reality Relevance} To ensure that the data distribution and tasks of LiveAgentBench are consistent with the real world, and to avoid inaccurate evaluation results caused by the gaps between the dataset and real-world tasks, the data of LiveAgentBench comes from public user cases on the Internet. Based on these data sources, we retained the capabilities and environments examined in the user cases through the standard process of SPDG. After in-depth analysis, we categorised these real-world questions into 104 specific scenarios. When designing and building the evaluation task, we must fully refer to the capabilities and scenarios of real user cases.

\textbf{Challenging} We have designed a standard process to identify user cases. Firstly, we should filter out questions that can be answered through simple searches and exclude these data from our data sources. Then, we choose the tasks to be solved with specific tools, such as browsers, to obtain real user cases with a certain level of challenge. Moreover, we invite experts to help us review and supplement our data.

\textbf{Ease of validation} When constructing the tasks, we require that the answers to the questions do not change over time. On the other hand, we modified the open questions into corresponding closed questions to ensure that the answers were sufficiently concise and unambiguous. When validating the agent's response, we use simple string processing, which is easy to handle and ensures the stability of the evaluation result.

\subsection{Social Perception-Driven Data Generation}

\begin{figure}[b]
  \centering
  \includegraphics[width=1\textwidth]{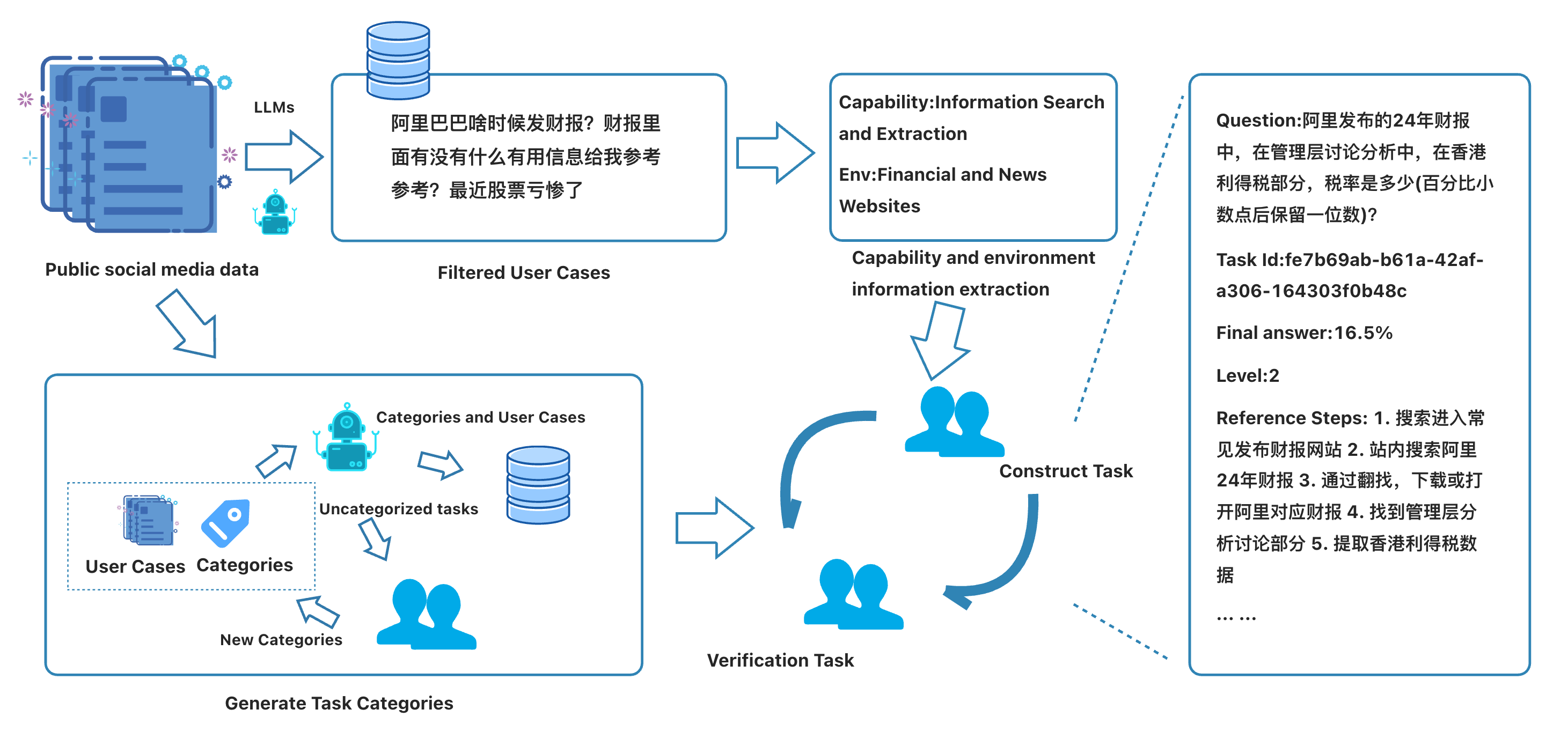}
  \caption{An illustrated introduction to the SPDG process, introducing the key aspects of the SPDG process by using a specific task as an example.}
\end{figure}

Social Perception-Driven Data Generation (SPDG) is an execution process for sustainable dataset production, which provides a standardised collaboration framework for human-machine. We have established a data production process with an operational specification system through systematically integrating human expertise and LLMs' capabilities. To be more specific, for the operational ambiguities in the data production, we define specific operational standards, such as problem reference standards, task production standards and a quality control mechanism. These standards ensure the quality of the output dataset. At the same time, we replaced part of the manual work with LLMs to improve the efficiency and consistency of the LiveAgentBench dataset. By doing this, iterations of LiveAgentBench are able to keep up with the changes in user needs and avoid inaccurate evaluation results caused by data contamination.

\subsubsection{Data Collection}
To ensure the realistic relevance of the data, based on user behavioural characteristics, we selected some representative Internet platforms as our data sources. Real user data is systematically collected from multi-source open platforms via an automatic and manual labelling combined process. The collected data sources are as follows:

\quad$\bullet$ \textbf{\textit{Open Q\&A communities}}: specific domain questions of various platforms such as Zhihu, Quora, and Baidu Knows.

\quad$\bullet$ \textbf{\textit{User-generated content platforms}}: comments and topics discussed on social media such as Xiaohongshu, BiliBili, and Douyin.

\quad$\bullet$ \textbf{\textit{Professional forums}}: posts and articles in technical communities 
such as Stack Overflow and CSDN.

\quad$\bullet$  \textbf{\textit{Video interaction platforms}}: Q\&A pop-up data from short video platforms, including TikTok and Kuaishou.

We have processed the collected data preliminarily to ensure the data sources are real and rich. Furthermore, we use LLMs to help screen out the cases with attachments to ensure the user cases carrying graphics, audio and video can not be ignored.

\subsubsection{Data Screening}

In the initial screening of Internet data sources, we require user cases to satisfy the feature of non-retrievability and tool dependency to ensure a certain level of dataset complexity. The user questions obtained after the initial screening fulfil our two basic requirements for the dataset: LLMs are not able to get the answer directly through simple retrieval-augmented generation (RAG), and they cannot answer the user question without at least one tool use. In this section, we have selected a total of 1112 user cases that meet the criteria. The definitions of non-retrievability and tool dependency are given below.

\textbf{\textit{Non-retrievability}} We filter purely knowledge-based user questions, thus, the dataset does not contain these cases. To further ensure the complexity of evaluation tasks, we also required that the user cases should not be directly answered through retrieval-augmented generation. This procedure ensures that all tasks in the dataset cannot be answered directly by simple thinking or searching.

\textbf{\textit{Tool dependency}} In order to conveniently and accurately filter out user cases with tool-dependent characteristics, we use LLMs as our annotators. When screening, we give LLMs a few-shot prompt to make them have a certain ability of tool judgment, and analyse the possible execution steps and tools to be used from our data sources. Based on this, we filter out the user tasks that have the characteristic of using tools from a large amount of corpus.



\subsubsection{Task Construction}


\textbf{\textit{Capabilities and Environment Extraction}} Firstly, we use a large model to generate possible execution steps for user use cases, and extract the required capabilities and environmental information based on these execution steps. As in the steps, it is necessary to access data from government websites and derive a conclusion through reasoning. It is considered that the ability required for this step is reasoning ability, and the environmental information is government websites and browsers.

\textbf{\textit{Task production}} Based on the environmental information and examined capabilities, we select proper annotators that are relevant to the specific category to build questions and labels. If there are no user tasks in the category, the annotators will produce the task based on their own life experiences and background knowledge. To ensure the complexity of the task, annotators are required to label the correct steps for the task execution, which is used for judging the complexity of the task. Moreover, because most user questions are open-ended, annotators will modify the query to make sure that the task's label is stable, sufficiently concise and unambiguous, which is important to ensure the evaluation results are easily verifiable.

\subsubsection{Quality Control}

During the task production process, we have designed corresponding standards to ensure the controllability of every step. Nevertheless, quality control of the whole process is still indispensable. To this end, we have designed manual and LLMs double-check mechanisms at several steps in the process, including task relevance checking, task complexity checking, planning executability checking, and result uniqueness checking.

\textbf{\textit{Task relevance checking}} During the process of task production, annotators may modify the user questions if the answer is open-ended, so it is necessary to check the relevance between the new and original tasks. We conduct the same procedures as capabilities and environment extraction to obtain the information of new tasks, and compare with the original tasks. If there is more than a 50\% mismatch between the environment and examined capabilities, the tasks will be filtered and reconstructed.

\textbf{\textit{Task complexity checking}} To accurately measure the task difficulty, we review our dataset through LLMs and pick out the cases where the labelled planning step is problematic, such as missing and invalid steps, and fix these problems. Referring to the definition of different levels from GAIA\citep{GAIA}, the difficulty of the task is classified based on the number of planning steps and the tools used. If the planning steps are fewer than 2 or there is no need to use tools, the task will be deleted from our dataset..

\textbf{\textit{Planning executability checking}} After labelling all the planning steps, we invite other annotators to conduct every step to verify that the correct answer can be obtained through the labelling steps.

\textbf{\textit{Result uniqueness checking}} Even though the uniqueness of answers was required during the task production, it is still impossible to avoid the existence of multiple answers for some tasks. In this respect, we conduct a double-blind annotation, where two respondents will participate in answering the task without any reference. If the answers are inconsistent, a third respondent will be involved to check whether the task is ambiguous and give the final answer. 

\section{Experiment}

We selected open-source and closed-source LLMs and autonomous agents that have achieved excellent performance in various benchmarks for a comprehensive evaluation. We report the overall performance primarily on LiveAgentBench, and then conduct a detailed analysis based on the scenarios and examined capabilities of the tasks.

\subsection{Experimental Setup}

Considering the capabilities of LLMs, we selected multimodal models, reasoning models, and other LLMs that are popular recently. On the other hand, to explore whether there is a difference between LLMs released by different companies in various cultural backgrounds, 5 LLMs released by American and Chinese were chosen, including Qwen3-235B-A22B\citep{Qwen3}, Claude35-sonnet\citep{anthropic_claude_35}, GPT-4o\citep{GPT-4o}, Gemini-2.5-pro\citep{Gemini_2.5} and Deepseek-R1-671B\citep{DeepSeek}.

Furthermore, we chose 4 agents released this year. Some of them are skilled at doing research, such as OpenAI Deep Research-4omini\citep{OpenAI_Deep_Research}, Perplexity-Research\citep{Perplexity_Deep_Research}, while some of them are familiar with using tools, including Manus\citep{Manus}, and Coze Space\citep{Coze_Space}. Additionally, we also used AWorld\citep{AWorld2025}, an open-source agent framework with Claude35-sonnet\citep{anthropic_claude_35} as the planning and execution model for our evaluation.

We evaluated all LLMs and agents only using their own capabilities and inner tools. If LLMs lack some capabilities, such as uploading attachments, then they will be directly considered to be failed in solving these problems. As for autonomous agents, we evaluate them directly on their official websites. In the actual evaluation process, LLMs is evaluated utilising the zero-shot prompting, and all the performance is evaluated using the Pass@1 metric.

\subsection{Answer extraction}

We designed the task results to be closed and unambiguous at the beginning of the data collection. Therefore, it is sufficient to use string matching to extract answers from the responses of LLMs and agents, and then determine whether the answer is correct or not. This method makes it easy and accurate to use when evaluating, and researchers can use it without involving any LLM as the judge model.

\section{Result and Analysis}
\subsection{Overall Performance}

\begin{table}[h]\small
    \caption{Overall performance of LLMs, agents and humans on LiveAgentBench."W\&S" represents Work and Study, "DL" represents Daily Life, "IA\&P" represents Information Access and Processing, "H\&SS" represents Humanities and Social Science, and "SP" represents Social Production. Overall represents the percentage of correctly solved problems by models or agents across all tasks, and scores in subcategories are the percentage of correctly solved problems within all tasks under the specific category. All of the scores are shown in the percentile system.}
    \label{Performances}
    \centering
    \renewcommand\arraystretch{1.2}
    \tabcolsep=0.12cm
    \begin{tabular}{ccccccc|cccc}
         \toprule
            \multirow{2}{*}{\textbf{Subject}} & \multirow{2}{*}{\textbf{Overall}} & \multicolumn{5}{c|}{\textbf{Scenario}} & \multicolumn{4}{c}{\textbf{Capability}} \\
            \cline{3-11}
            & & W\&S & DL & IA\&P & H\&SS &  SP & Text File & Image & Video & Audio \\
            \midrule
            \rowcolor{gray!8}
            \multicolumn{11}{c}{LLMs} \\
            \midrule
            Qwen3-235B-A22B & 7.75 & 16.39 & 8.25 & 6.38 & 3.61 & 6.17 & 8.02 & 0 & 0 & 0 \\
            Claude35-sonnet & 8.28 & 13.11 & 9.28 & 8.51 & 4.82 & 7.41 & 6.13 & 15.13 & 0 & 0 \\
            GPT-4o & 9.09 & 13.11 & 11.34 & 4.26 & 6.02 & 9.88 & 5.19 & 19.33 & 0 & 0 \\
            Gemini-2.5-pro & \textbf{16.85} & 19.67 & \textbf{18.56} & \textbf{12.77} & \textbf{19.28} & \textbf{13.58} & 12.26 & 27.73 & 16.0 & 0 \\
            Deepseek-R1 & 9.89 & \textbf{21.31} & 6.19 & 6.38 & 8.43 & 9.88 & 13.2 & 0 & 0 & 0\\
            \midrule
            \rowcolor{gray!8}
            \multicolumn{11}{c}{Agents} \\
            \midrule
            Gemini Deep Research & 14.17 & 11.48 & 12.37 & 19.15 & 10.84 & 17.28 & 24.3 & 0 & 0 & 0 \\
            Manus & \textbf{35.29} & \textbf{40.98} & \textbf{31.18} & \textbf{40.42} & \textbf{39.76} & \textbf{28.40} & 37.85 & 35.29 & 16.0 & 33.33 \\
            OpenAI Deep Research & 27.54 & 19.67 & 28.87 & 38.30 & 20.48 & 25.93 & 33.49 & 24.17 & 4.0 & 13.33 \\
            Perplexity Research & 23.80 & 26.23 & 25.77 & 29.79 & 24.10 & 13.58 & 30.95 & 20.17 & 0 & 0 \\
            Coze Space & 18.45 & 19.67 & 19.59 & 19.15 & 15.66 & 17.28 & 25.23 & 10.08 & 0 & 13.33 \\
            \midrule
            \rowcolor{gray!8}
            \multicolumn{11}{c}{Framework} \\
            \midrule
            AWorld & \textbf{15.51} & 21.31 & 13.40 & 12.77 & 16.87 & 14.82 & 13.81 & 19.33 & 16.0 & 13.33 \\
            \midrule
            \rowcolor{gray!8}
            \multicolumn{11}{c}{Human} \\
            \midrule
            Human & \textbf{69.25} & 75.41 & 74.23 & 74.47 & 62.65 & 64.20 & 73.33 & 60.50 & 80.0 & 73.33 \\
         \bottomrule
    \end{tabular}{}
\end{table}

The evaluation results on LiveAgentBench are in Table \ref{Performances}. Based on the results, we can see that there is still a significant margin for improvement for LLMs and agents when completing real-world tasks. LLMs can only complete approximately 13.48\% of the missions in LiveAgentBench, while the agents perform relatively better. The integration of tools enables agents to obtain more accurate information, thus further expanding the capability boundaries of them. However, despite being equipped with ordinary tools, agents still complete only 23.85\% of the tasks. In contrast, humans can complete approximately 69.25\% of the tasks without effort.

\subsection{Specific Insights}

Overall, the performance of LLMs lags behind that of agents and the agent framework on LiveAgentBench. Based on the table, a conclusion can be drawn as follows. The lack of tools leads to a significant disparity in performance between LLMs and agents. The overall score of LLMs is approximately 56.51\% weaker than that of autonomous agents on average. Different from the other results, the performance of Genimin-2.5-pro is better than Gemini Deep Research with 2.5pro, and the two scores are 16.85 and 14.17, respectively. The main reason is that Gemini Deep Research does not yet support image, audio and video uploads, which leads to a lower score. However, in terms of text file processing, Gemini Deep Research scored 24.3, significantly higher than Genimin-2.5-pro. It demonstrates that increased tool capabilities can substantially improve the LLM's ability to solve real-world problems.

Furthermore, from the results in different scenarios, we observe that there is a large gap between LLMs' performance in the two categories of Work and Study scenario and Information access and processing, while agents perform comparably in these two scenarios. In the field of Work and Study, most information comes from attachments provided by the task, whereas browser utilise is required to solve problems in the scenario of Information access and processing. The lack of tools leads to a substantial disparity between LLMs.

Although agents perform better than the base model on LiveAgentBench, there is still a significant gap compared to humans. Manus, the highest-scoring agent, has already supported most daily-use tools, but there is still a 33.96-point difference compared to humans. During the evaluation, we identified two main factors causing agents to fail to complete the task. Firstly, the instability prevented agents from obtaining accurate information effectively. Secondly, a lack of environmental background knowledge leads to the agent not being able to locate information in unfamiliar environments.

Finally, there is still a gap in performance between the open-source agent framework and agents, mainly due to the framework's stability lagging behind that of agent products. During the evaluation of the AWorld framework, we found that 11.76\% of task failures were caused by this reason, including the instability of inner tools. 

\subsection{Error Analysis}

To further investigate the failures caused by tool instability and the lack of environmental background knowledge, we have sampled a number of failure cases and provided them in the Appendix. In these cases, issues such as the website shutdown and audio/video reading failures caused by the tools during execution prevent agents from obtaining valid information. As a result, the task failed directly, despite agents having planned the solving steps correctly. In contrast, when humans perform tasks, the tool instability rarely happens, which reveals that there is a significant difference between agents' execution environment and that of humans. To further improve agent performance, enhancing the stability of the task execution environment is necessary.

In specific cases of missing environmental background knowledge, we found that agents can not find the required information efficiently after locating the correct websites or other tools. The root reason is that different websites or tools - what we call different environments - have different functions, layouts and logics. For example, government and commercial websites differ in structured information, while music and video tools' functional layouts are completely different. Without background knowledge about such websites and tools, it becomes extremely difficult for agents to locate the required information accurately in unfamiliar environments. In a specific failure case, when entering a relatively unfamiliar website (such as a government website), the agent can not find the sub-entry for the required data.

\section{Conclusion}

In this work, we propose the LiveAgentBench, a comprehensive benchmark that comprises 104 scenarios and aligns with the distribution of real-world problems. Additionally, we propose a social perception-driven data generation process (SPDG), based on which we can update LiveAgentBench in a standard and efficient way, avoiding data contamination and keeping pace with the needs of real-world users. To sum up, LiveAgentBench aims to facilitate the detailed evaluation and understanding of general AI agents, driving further advancements in this field.

\textbf{Limitations} Currently, we are mainly focusing on real-world tasks evaluation in Chinese, so LiveAgentBench is predominantly composed of Chinese-language tasks and lacks cultural diversity. But we will use SPDG to collect corpus from online sources and continuously update the datasets to ensure the diversity of LiveAgentBench in our further work. One of the core principles of LiveAgentBench is to stay close to reality, but this proximity implies that tasks may be open-ended and contain ambiguities. To balance these two aspects, we have invested significant effort in disambiguation and modifying the data, which has resulted in some unnatural details in the tasks. Although these processes ensure that each task has only one correct answer, they also make the tasks inconsistent with human habits. We will continue to optimise and improve our overall processes to address these issues in future work.

\bibliographystyle{rusnat}
\bibliography{reference}

\end{document}